\documentclass[letterpaper, 10 pt, journal, twoside]{IEEEtran}
\IEEEoverridecommandlockouts

\usepackage{cite}
\usepackage{amsmath,amssymb,amsfonts}
\usepackage{fancyhdr}
\usepackage{algorithmic}
\usepackage{graphicx}
\usepackage{textcomp}
\usepackage{float}
\usepackage{dblfloatfix}
\usepackage{hyperref}
\usepackage{cleveref}
\usepackage{caption}
\usepackage{float}
\usepackage{cuted}
\usepackage[table]{xcolor}

\newcommand{\edit}[1]{{{#1}}}
\newcommand{\mc}[1]{\mathcal{#1}}
\newcommand{\firstplace}{\cellcolor{green!25}}
\newcommand{\secondplace}{\cellcolor{yellow!25}}
\title{HAMMER: Heterogeneous, Multi-Robot \\ Semantic Gaussian Splatting}

\author{Javier Yu$^{1}$, Timothy Chen$^{1}$, and Mac Schwager$^{1}$
\thanks{This work was supported in part by DARPA grant HR001120C0107, ONR grant N00014-23-1-2354, and MIT-Lincoln Labs grant 7000603941.  The second author was supported by the NASA NSTGRO fellowship.}
\thanks{$^{1}$Javier Yu, Timothy Chen, and Mac Schwager are with the Department of Aeronautics and Astronautics, Stanford University, Stanford, CA 94305, USA {\tt\footnotesize \{javieryu, chengine, schwager\}@stanford.edu}}%
\thanks{DOI: 10.1109/LRA.2025.3575235 \copyright 2025 IEEE.  Personal use of this material is permitted.  Permission from IEEE must be obtained for all other uses, in any current or future media, including reprinting/republishing this material for advertising or promotional purposes, creating new collective works, for resale or redistribution to servers or lists, or reuse of any copyrighted component of this work in other works.}
}

\begin{document}


\maketitle

\begin{strip}
\begin{minipage}{\textwidth}\centering
\vspace{-95pt}
\includegraphics[width=0.65\textwidth]{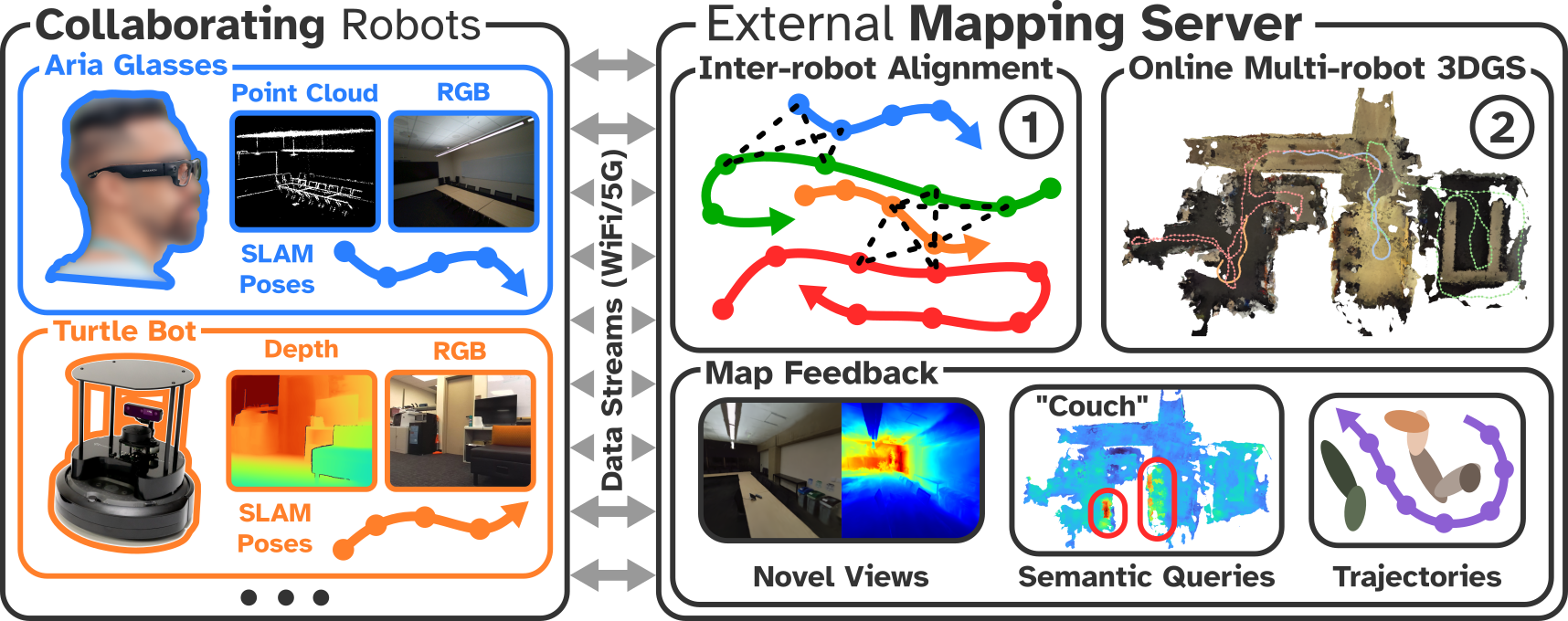}
\captionof{figure}{HAMMER takes streaming image and depth data from a heterogeneous team of robots and edge devices (e.g. Aria Glasses \cite{engel2023project}), each running on-board SLAM in unaligned local frames.  \edit{On a GPU server, HAMMER aligns new device data streams into a global frame in a one-time computation (1)}. All aligned data streams are then integrated into an online continually trained semantic $3$DGS map (2). The map can be leveraged for open-vocabulary object localization and planning.}
\label{fig:abstract}
\vspace{-10pt}
\end{minipage}
\end{strip}

\begin{abstract}
3D Gaussian Splatting offers expressive scene reconstruction and can model a broad range of visual, geometric, and semantic information. However, efficient real-time map reconstruction with data streamed from multiple robots and devices remains a challenge. To that end, we propose HAMMER, a server-based multi-robot Gaussian Splatting method that leverages ROS communication infrastructure to generate 3D, metric-semantic maps from asynchronous robot data-streams. HAMMER consists of (i) a \edit{one-time} frame alignment module that transforms local SLAM poses and image data into a global frame and requires no prior relative pose knowledge, and (ii) an online module for \edit{continually} training semantic 3DGS maps from streaming data. HAMMER handles mixed perception modes, adjusts automatically for variations in image pre-processing among different devices, and distills CLIP semantic codes into the 3D scene for language queries. In real-world experiments, HAMMER creates better maps compared to baselines and is useful for downstream tasks, such as semantic navigation (e.g., ``go to the couch"). Accompanying content at \href{hammer-project.github.io}{hammer-project.github.io}.
\end{abstract}


\section{Introduction}
\label{sec:intro}
\IEEEPARstart{M}{apping} is a fundamental component in the autonomy stack. Recent advances in scene modeling with differentiable rendering \cite{mildenhall2021nerf, kerbl20233d} have enabled robots to generate high-fidelity scene reconstructions in real time. Compared to traditional methods that use meshes, point clouds, or voxel grids to model physical features of the environment, these newer methods reason about the scene as a radiance field, represented as a  parametric model regressed from photometric rendering losses. Radiance field-based methods can achieve photorealistic novel-view synthesis and even capture thin, transparent, and reflective surface geometries that are challenging to model with traditional representations. Furthermore, radiance fields can be augmented with additional information channels---like semantic embeddings from a pre-trained vision-language models---making them well-suited for language-driven robotics applications. In particular, 3D Gaussian Splatting \cite{kerbl20233d} (3DGS) has been applied to a wide range of robot tasks including SLAM \cite{matsuki2024gaussian}, safe teleoperation \cite{chen2024safer}, manipulation \cite{shorinwa2024splat}, and navigation \cite{chen2024splat}.

Multi-robot mapping is useful for rapidly exploring new environments, but when combined with traditional $3$D reconstruction methods, can be difficult to scale efficiently, especially for teams of  heterogeneous robots that have a mix of mobility and sensing capabilities. Alternatively, 3DGS is a promising representation for multi-robot mapping because of its scalability to large environments \cite{xiong2024gauu}, modeling fidelity, and generalization to a broad range of sensing modalities including RGB, stereo depth, LiDAR \cite{zhou2024drivinggaussian}, and semantics \cite{yu2024language}.

In this work, we propose HAMMER, \underline{H}eterogeneous \underline{A}synchronous \underline{M}ulti-robot \underline{M}apping of \underline{E}nvironmental \underline{R}adiance. HAMMER enables a server communicating with a team of robots to construct a joint 3DGS map of an unknown environment. First, robots broadcast color images and depth maps, annotated with poses using the robot's on-board localization system, to a mapping server. \edit{The server extracts semantic information from the images, aligns the local poses of each robot to a global map frame, and continually trains in real-time a metric-semantic $3$DGS map using the robot data streams.} 

A server-based architecture allows our method to be used with existing robot and edge device hardware without high-powered GPUs, while leveraging typical communication infrastructure (e.g. WiFi/5G) to offload demanding frame alignment and 3DGS training to a central GPU server. We intentionally propose this partially centralized architecture to ensure that HAMMER is real-time executable with existing robots, wearable devices, and WiFi/5G communication infrastructure.  Computing requirements for 3DGS training are currently beyond the on-board compute capabilities of most robots and wearables.  When on-board computing matches the power of current high-end GPUs, fully distributed algorithms \cite{yu2022dinno,shorinwa_distributed_2023-1} may become a compelling alternative. 

HAMMER is designed to generalize to a wide range of robots and devices, combining the advantages of each device into a single map. A shared map enables these robots to have comprehensive spatial awareness compared to their own local maps. For example, drones can cover blind spots of mobile ground robots performing semantically-guided navigation. 

\edit{The contributions of HAMMER (Fig. \ref{fig:abstract}) are (1) a robust and metric inter-robot frame alignment using RGB images to align coordinate systems across devices with different image sensors and SLAM algorithms, and (2) an online $3$DGS training scheme from heterogeneous, asynchronous devices that produce posed RGB images with depth or point clouds. The frame alignment (1) occurs only once when a new robot joins the map server, while the real time training (2) operates continually, integrating aligned robots' data into the server map as it is streamed. Importantly, HAMMER does not require any prior knowledge on the initial robot poses nor inter-robot observations, enabling asynchronous and ad-hoc deployments.} 

\edit{We compare HAMMER against state-of-the-art multi-robot mapping baselines \cite{yugay2024magic, hu2023cp} across four scenes from the Replica dataset\cite{straub2019replica}, where HAMMER outperforms or matches the baseline while requiring less than a tenth the computation time.} Furthermore, in hardware deployments with 3-4 devices across two real-world environments, HAMMER provides more than 40\% better Mean-Squared-Error \edit{for novel-view image reconstruction}, compared to \cite{asadi2024di}.

\section{Related Work}
\label{sec:related}
\textbf{Radiance Fields and Gaussian Splatting.}
NeRFs, proposed in \cite{mildenhall2021nerf}, use neural networks to model radiance fields from posed RGB images and demonstrate photorealistic novel view synthesis and high-fidelity geometric reconstruction by optimizing these networks using a photometric loss with a differentiable rendering process. Follow up work with optimized model architectures \cite{muller2022instant} greatly reduced NeRF training times. However, NeRFs remain computationally expensive to optimize and query. As an implicit representation, NeRFs make it difficult to extract the underlying scene geometry. On the other hand, $3$DGS \cite{kerbl20233d} maintains many of the benefits of NeRFs and is significantly faster to optimize. The use of geometric primitives as a representation also facilitates collision and contact modeling with the environment \cite{chen2024splat}.

\textbf{Online 3DGS Training and 3DGS-based SLAM.}
$3$DGS is typically optimized using the following offline optimization scheme: (i) images are captured using a high-resolution camera, (ii) poses are estimated using structure-from-motion (SfM) like COLMAP \cite{schoenberger2016sfm}, and (iii) a $3$DGS map is optimized on the static dataset. However, this processing and training scheme is not well suited for robotics tasks where data is incrementally received from onboard sensors and prevents the $3$DGS map from being used during deployment. LEGS \cite{yu2024language} proposes using an incrementally optimized semantic 3DGS map with camera poses initialized from an on-board SLAM system. Both \cite{matsuki2024gaussian} and  \cite{keetha2024splatam} go a step further, using $3$DGS as the map representation for SLAM.

\textbf{Multi-robot Mapping.}
Mapping using multi-robot systems is a widely studied field, and many diverse methods have been proposed. Historically, multi-robot mapping and SLAM systems have utilized more traditional robot map representations like point clouds \cite{huang2021disco, hu2023cp} and meshes \cite{tian2022kimera}. Collaborative mapping is well-suited for scenarios requiring robust and rapid exploration \cite{chang2022lamp}. MACIM \cite{deng2024macim} proposes a decentralized multi-robot mapping algorithm that optimizes a signed distance function, while other approaches use NeRFs \cite{zhao2024distributed,asadi2024di}. Of these works, only Di-NeRF \cite{asadi2024di} addresses the problem of inter-robot frame alignment, which it does through gradient-based optimization. 

\edit{Developed concurrently to this work, MAGiC-SLAM \cite{yugay2024magic} proposes a multi-robot $3$DGS SLAM pipeline comprised of robots running independent $3$DGS SLAM instances whose individual maps are fused into a global map. MAGiC-SLAM requires significantly greater computational demands compared to HAMMER. Either each robot is equipped with a GPU capable of 3DGS optimization or data is streamed to a server with $N+1$ GPUs, where $N$ is the number of concurrently operating robots. Instead, HAMMER only requires a server with a single GPU and does not impose additional compute constraints on the robots beyond requiring a stream of pose and image/depth data. HAMMER offers similar or better reconstruction quality at one tenth the computation time, is built on ROS for real robots and on-line continual mapping, and embeds semantic codes for language based map queries.}

\section{Method}
\label{sec:mrplat_methods}
\subsection{Problem Setup}
\label{sec:setup}
We consider a set of $N$ robots, indexed $i \in \{1, \dots, N\}$, deployed in a shared environment. Their deployments can be asynchronous, potentially lacking any temporal overlap between individual robot deployments. Each robot produces color images, geometric information (e.g. depth images or point clouds), and camera pose estimates in $SE(3)$ with respect to an arbitrary local coordinate frame $\mc{T}^i$. These estimates can be retrieved from onboard VIO or SLAM commonly available on commercial robots and devices. The robots continuously stream data to a mapping server during deployment. The data is received incrementally by the server as the robot navigates through the environment. The goal is to continuously incorporate every robot's data into the $3$DGS map within a consistent global coordinate frame $\mc{T}^g$. We assume that there is sufficient visual overlap within the data streams of the robots such that alignment is possible. \edit{We also assume that at least one of the robots provides pose estimates in a metric coordinate frame.}

\subsection{Inter-robot Alignment}
\label{sec:alignment}
\begin{figure*}[ht]
    \centering
    \vskip 5pt
    \includegraphics[width=0.9\linewidth]{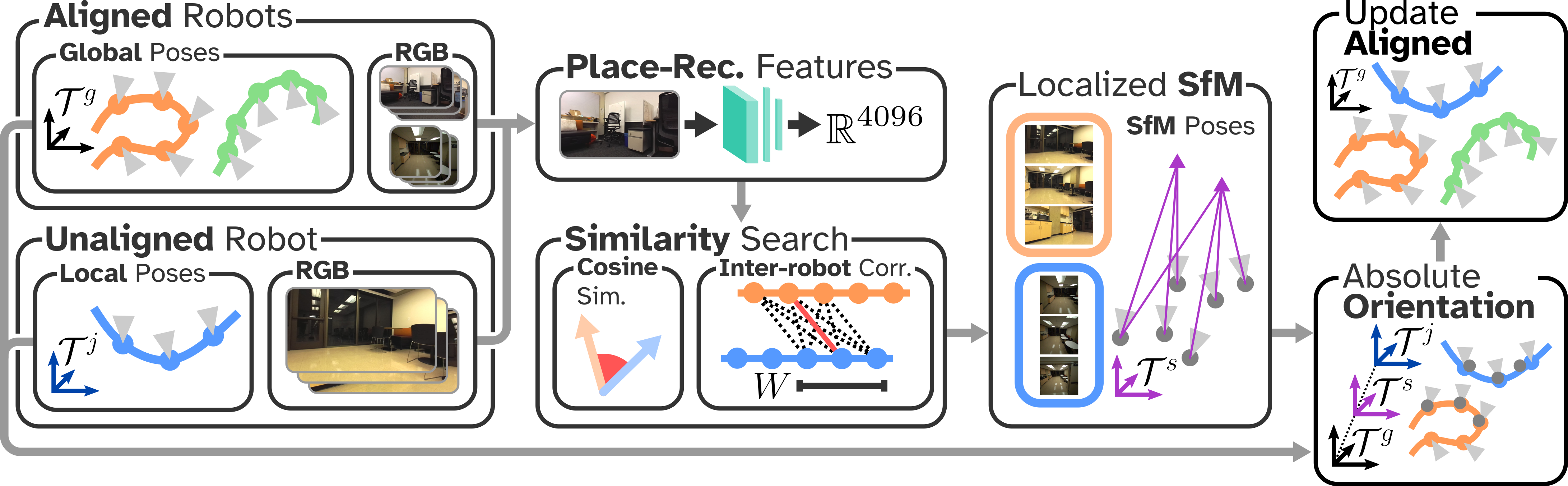}
    \caption{\edit{HAMMER uses a \emph{one-time} computation to align a new robot's data stream with the server's map frame. We match images from the unaligned robot with images from the server map using a place recognition feature extractor. When a correspondence is verified, matched images from robot and server are passed to an SfM solver, from which we find a single transform to align the robot to the server map frame. From then on, images and transformed SLAM poses from the robot are streamed into the data loader where they are continually refined as a part of $3$DGS training}.}
    \label{fig:align}
    \vspace{-10pt}
\end{figure*}

\edit{When a new robot or device joins the server, HAMMER performs a \emph{one-time} frame alignment to bring the new robot's data stream into alignment with the server's map frame (Fig. \ref{fig:align}).  This computation takes approximately $36$ sec. in our implementation.  During this time, the robot's data stream is cached at the server and integrated into the map after an alignment is found.  From then on, the robot's data is continually fused into the server's $3$DGS map in real time.}

\subsubsection{Correspondences}
First, HAMMER detects potential candidate correspondences between the data of the unaligned robots and that of the aligned robots. To succinctly describe the content of images, a place-recognition feature vector is extracted for every color image received from the collaborating robots using NETVLAD \cite{arandjelovic2016netvlad}. The place features from the unaligned robots are periodically compared to those of the aligned robots using cosine similarity. For a pair of images derived from aligned robot $i$ and unaligned robot $j$ with similarities exceeding $\gamma_{ij}$, an additional feature-matching verification step is performed where per-image SuperPoint features \cite{detone2018superpoint} are extracted and matched using LightGlue \cite{lindenberger2023lightglue}. If the fraction of matched features exceeds a fixed ratio $\xi = 0.25$ then the image pair is accepted as a potential inter-robot correspondence.

\subsubsection{Adaptive Thresholding} 
When no image pairs exceed $\xi$, then $\gamma_{ij}$ is increased to match the mean of all place recognition similarities greater than the previous value. Adaptive thresholding helps combat unilaterally higher similarities of images captured from the same camera on different robots regardless of visual content. $\gamma_{ij}$ is initially set to a low value for all robot pairs, circumventing any hyper-parameter tuning that would be necessary. While determining candidate alignment image pairs grows quadratically in the number of images per robot, many of these operations can be cached, forming a relatively small computational footprint.

\subsubsection{Localized Structure-from-Motion}
For every potential correspondence between an aligned and unaligned robot, a small set of color images within some fixed temporal window size $W = 16$ is selected from the aligned and unaligned robot's data cache on the server (i.e. $32$ images). These images are used to perform \emph{localized} SfM. This procedure is in contrast to \emph{bulk} SfM, where the entirety of the robot's data stream is used. To perform SfM, we use the COLMAP backend \cite{schoenberger2016sfm} with SuperPoint features and the SuperGlue matcher \cite{sarlin2020superglue}, which have exhibited robustness in aligning images from heterogeneous devices. Importantly, the localized SfM does not incorporate any pose information from the onboard SLAM systems, producing an independent relative pose estimate between robots.

\subsubsection{Registration}
The output of the localized SfM procedure in $\mathcal{T}^{s}$ provides a transformation link from the unaligned robot's SLAM frame $\mathcal{T}^j$ to the aligned robot's world frame $\mathcal{T}^{g,i}$, which by definition is also the global frame of all other aligned robots $\mathcal{T}^g$. Computing the relative transformation from $\mathcal{T}^j \to \mathcal{T}^s$ and $\mathcal{T}^s \to \mathcal{T}^g$ requires solving two instances of a modified absolute orientation problem \cite{umeyama1991least} for scale, rotation, and translation. The first instance computes the transformation from unaligned robot $j$'s $W$ SLAM poses in $\mathcal{T}^j$ to the corresponding $W$ SfM poses in $\mathcal{T}^s$. The other transforms aligned robot $i$'s $W$ SfM solutions in $\mathcal{T}^s$ to its corresponding aligned $W$ poses in global frame $\mathcal{T}^g$.
To compute these transformations, we solve the following \emph{rotation-aware} absolute orientation problem
\begin{equation}
\label{eq:srt_optimization}
    \min_{s, R, t} \sum_{k=1}^W \left[ ||s R \, t_k^j - t_k^s + t||^2 + \epsilon ||R R_k^{j} - R_k^{s}||_F \right],
\end{equation}
where $||\cdot||_F$ denotes the Frobenius norm. \Cref{eq:srt_optimization} optimizes the scaling, rotation, and translation $(s, R, t)$ between the two frames with a small regularization term on the rotation to address degenerate data. 
While no closed form solution exists for (\ref{eq:srt_optimization}), each individual term has a closed form solution. Therefore, for small $\epsilon$, an approximately optimal solution can be computed. 

The composition of both relative transformations yields $(s^{jg}, R^{jg}, t^{jg})$, the relative frame transformation $\mathcal{T}^j \to \mathcal{T}^g$. \edit{Importantly, given that the origin robot estimates poses in metric scaling, all $s^{jg}$ scales to a metric coordinate frame $\mathcal{T}^{g, 0}$. Therefore, other than the origin robot, no other robots need to run metric SLAM methods.}

\edit{The proposed inter-robot alignment module is robust and fails gracefully by terminating early or rejecting spurious alignments. During runtime, HAMMER rejects alignments where the localized SfM fails to estimate poses for all $2W$ input images or alignments that have high translation ($0.1$m in the map frame) or rotation errors ($10\deg$) as computed by (\ref{eq:srt_optimization}). We also emphasize that the alignment process is performed only once for each robot with the exception of the origin robot which does not require alignment. Specifically, upon the first successful alignment, robot $j$ is immediately designated as ``aligned" and no longer performs any alignment procedure even if there are available potential correspondences. Therefore, although alignment is costly, it is a one-time cost to produce the transform $\mathcal{T}^j \to \mathcal{T}^g$.}

\Cref{fig:align} illustrates a high-level overview of the alignment process. Importantly, the process treats the onboard localization algorithms as black-boxes and only uses the resulting camera poses and color images as input. This attribute of HAMMER dramatically simplifies the integration of new devices due to the ubiquity and variety of on-board localization frameworks. In contrast, alignment methods that require access to lower-level information from each individual robot's localization system (e.g. pose graphs or sparse maps) will fail if the software is closed-source.

\subsection{Multi-robot 3DGS Mapping}
\label{sec:mapping}
A naive approach to multi-robot map building is to collect all robot data into a large dataset, align the data through \emph{bulk} SfM, and optimize a $3$DGS model from the static dataset. This procedure is similar to how neural fields are conventionally trained. However, offline optimization of the map prevents the robot team from using the map \emph{during} deployment and on-demand, reducing the map's effectiveness.

Instead, HAMMER continually trains the global map from steaming robot data, \edit{adding new robots and devices to the set of contributing robots with the alignment module from Sec.~\ref{sec:alignment}}. To achieve this functionality, our $3$DGS mapping system performs online optimization on a stream of data and flexibly incorporates data from heterogeneous devices with possibly different camera sensors and geometric scene data (e.g. point clouds or depth images). When only RGB images are available, monocular depth estimation \cite{depthanything} could be used as geometric scene data if it produces metric depth. However, handling such cases is outside the scope of our implementation of HAMMER.

\begin{figure*}[]
    \vskip 5pt
    \centering
    \includegraphics[width=0.95\linewidth]{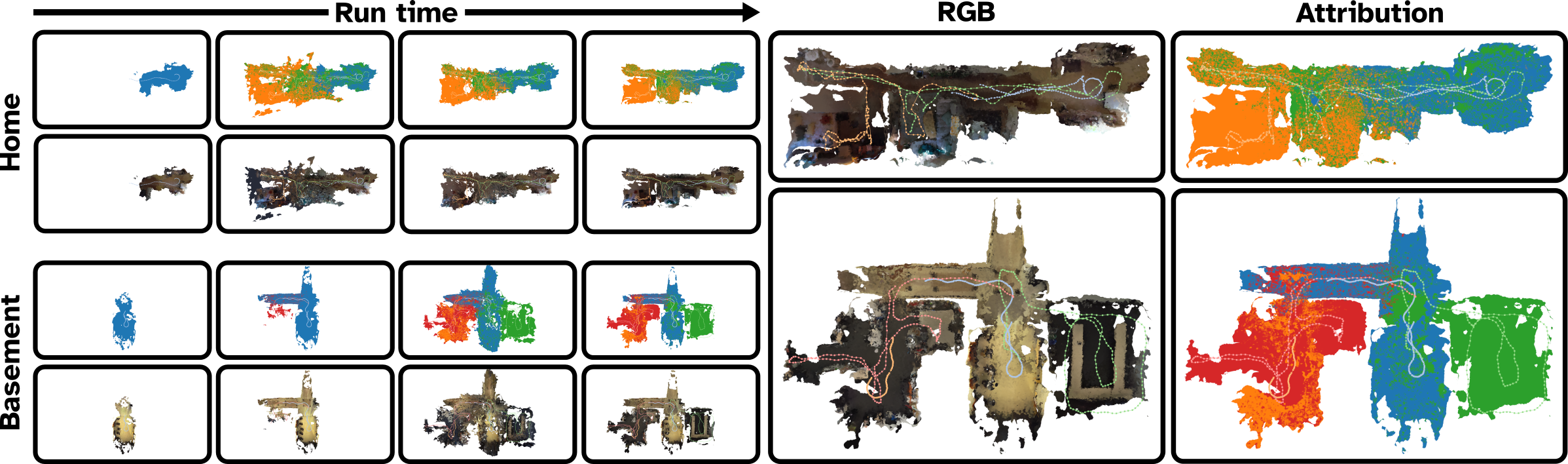}
    \caption{
    Meshes extracted from HAMMER's map during runtime (\emph{left}) and the final map reconstruction (\emph{right}). The attribution meshes show the part of the map contributed by each robot\edit{: Aria 1 (red), Aria 2 (green), GR 1 (orange), and GR 2 (blue)}.
    }
    \label{fig:mesh}
    \vspace{-10pt}
\end{figure*}

\subsubsection{Representation} 
3DGS models the opacity and color of the environment using explicit Gaussian primitives, which are optimized based on a differentiable, tile-based rasterization process into a photometric loss. The optimizable parameters of the $k$-th Gaussian include a mean $\mu_k \in \mathbb{R}^3$, a covariance $\Sigma_k \in S^3_{++}$ (the geometric extent of the Gaussian), an opacity $\alpha_k \in [0, 1]$, and an RGB color $c_k \in [0, 1]^3$. The color is further represented as a set of learned spherical harmonics coefficients, yielding view-dependent color to model realistic visual appearance.

Aside from color, HAMMER is also able to model semantics in the environment. Specifically, per-pixel semantic embeddings, extracted from color images using CLIP \cite{radford2021learning, zhou2022extract}, are distilled into the $3$DGS. \edit{The choice of pixel-wise semantic feature extractors is flexible, but we choose CLIP because of its strong zero-shot performance \cite{esmaeilpour2022zero}.} Some works \cite{qin2024langsplat} augment the parameters of each Gaussian with a semantic channel and regress it directly from an image reconstruction loss. However, these semantic embeddings can vary up to 1000 dimensions, which is prohibitively expensive to inference and store. Therefore, some methods store an encoding network used to compress the CLIP embedding. This encoding network is learned in conjunction with the $3$DGS, but can be difficult to optimize while streaming data. 

Alternatively, HAMMER optimizes a continuous feature field, parameterized by a neural network with hash-grid positional encodings. This feature field $\phi_s(x): \mathbb{R}^3 \rightarrow \mathbb{R}^{d_s}$ maps positions in the $3$DGS frame $\mathcal{T}^g$ to a feature vector of dimension $d_s$. To query the semantic embedding of a Gaussian, we evaluate $\phi_s$ at $\mu_k$. During training, we re-project the $3$DGS rendered depth and ground-truth CLIP images into $3$D, producing a labeled point cloud. The point cloud is then directly used to supervise the feature field. \edit{Gradient flow from the point cloud to the underlying model is blocked in order to simplify gradient computation as we do not observe improved scene reconstruction otherwise.}

A key challenge when mapping with heterogeneous data streams is compensation for each sensor's different image signal processing (ISP) pipelines (e.g. exposure and white-balancing). These pipelines can have a significant impact on the appearance of the resulting color images. Consequently, these appearance shifts can produce view inconsistency in the color between devices, despite their content being the same. The view inconsistency can cause erroneous, non-existent structures in the 3DGS map,  called \emph{floaters}, due to the optimizer reconciling contradicting signals from images  captured from devices with different ISP pipelines. To avoid this undesirable behavior, we utilize the bilateral grid-based ISP compensation method proposed in \cite{wang2024bilateral}. Specifically, for each image, we create additional bilateral grid parameters that compensate for the appearance differences between devices by matching the color of the rendered images to the ground-truth images through affine transformations.

\subsubsection{Online Optimization}
Upon successful alignment, robot $j$'s data in the cache and all future data can be incorporated into the 3DGS map. For each message sent by aligned robots, HAMMER initializes the Gaussian geometric parameters (i.e. $\mu_k, \Sigma_k$) based on the pose and geometric data (i.e. depth images or point clouds) and the color based on the corresponding pixel in the RGB image. Specifically, if a depth image is received, then the camera intrinsics and pose are used to project random pixels into 3D to create a sparse point cloud. Alternatively, if a point cloud is received, then a fixed number of points are sampled to produce a sparse point cloud. Next, new Gaussians are spawned, with means $\mu_k$ located at the points of the sparse point cloud.
The Gaussian covariances $\Sigma_k$ are initialized with isotropic covariances with pixel-level resolution. Mathematically, these covariances are computed as $\Sigma = \frac{d}{f} \*I$ where $d$ is the distance of the Gaussian from the source camera, $f$ is the focal length of that camera, and $\*I$ is identity.
Lastly, the opacity is initialized to $\sigma = 0.3$ for all Gaussians.

After the initialization phase, the data from each aligned robot stream is aggregated into a pool, which is continuously sampled and used for $3$DGS optimization. We weight the probability of a particular data tuple being sampled based on the number of times it has already been sampled to avoid over-weighting information from frames received early in the runtime. When depth images are available, HAMMER adds depth-supervision to the existing color photometric loss. The depth loss helps improve the geometry of the map in regions of visual ambiguity (e.g. monochromatic flat surfaces).

\subsubsection{Pose Refinement}
\label{sec:meths_pose}
\edit{Although the alignment module produces robust estimates of the local-to-world transforms, it cannot account for gradual drift or other temporal noise. To combat this deficiency, in addition to optimizing the scene parameters, HAMMER treats the SLAM pose and the local-to-world transform $\mathcal{T}^j \to \mathcal{T}^g$ as initial estimates that are continually optimized as part of $3$DGS training, incrementally deforming each agent's trajectory to create a visually consistent global map.} Specifically, we optimize an $SE(3)$ offset from the aligned camera pose to the refined camera pose. Regularization on both the rotation and translation is used to avoid catastrophic drift in the poses. We use a similar approach for fine-tuning $\mathcal{T}^j \to \mathcal{T}^g$.

\section{Experiments}
\label{sec:results}
\edit{First, we compare HAMMER to state-of-the-art baselines \cite{hu2023cp, yugay2024magic} by assessing their reconstruction accuracy on the \emph{ReplicaMultiAgent} dataset \cite{straub2019replica, hu2023cp}. However, ReplicaMultiAgent only contains scenes from simulated environments, and lacks heterogeneous robots/sensing devices and challenging real-world scene conditions (e.g. motion blur, diverse lighting). Therefore, to showcase the generalizability of HAMMER and its real-time deployment in real-world environments, we also assess its performance in two different hardware trials with data collected using real robots.}

\subsection{Experimental Setup}
\subsubsection{Implementation}
Unlike other baselines, HAMMER is fully integrated into the ROS2 ecosystem \cite{quigley2009ros} to stream data from all robots to a server, and uses a modified version of NerfBridge \cite{yu2023nerfbridge} for map building. The mapping server is a desktop computer with a NVIDIA RTX 4090 GPU. 5GHz WiFi is used to communicate during hardware trials.

\emph{ReplicaMultiAgent} is composed of two robots covering four scenes featuring mesh reconstructions of real indoor environments. Each robot produces RGBD images. To simulate real-time deployment within the environments, we create 2 minute long ROSBags that stream data at 20Hz. Each robot uses KISS-ICP \cite{vizzo2023kiss} to perform localization and streams this data simultaneously with the other robot.

For the hardware trials, the heterogeneous teams are composed of ground robots (GRs) equipped with ZED stereo cameras, and humans wearing Aria glasses \cite{engel2023project} that carry a range of onboard sensors. Onboard the GRs, RGBD images ($1280\text{x}720$) are published onto the network at 10 Hz along with poses estimated onboard using the ZED SDK. Aria glasses estimate pose using the Aria Project's proprietary SLAM system. The Arias produce $1440\text{x}1440$ fisheye RGB images and point clouds at 10 Hz. Deployments range between 1-2 minutes, and start offset by 20 seconds each.

\begin{figure*}[ht]
    \vskip 5pt
    \centering
    \includegraphics[width=0.95\textwidth]{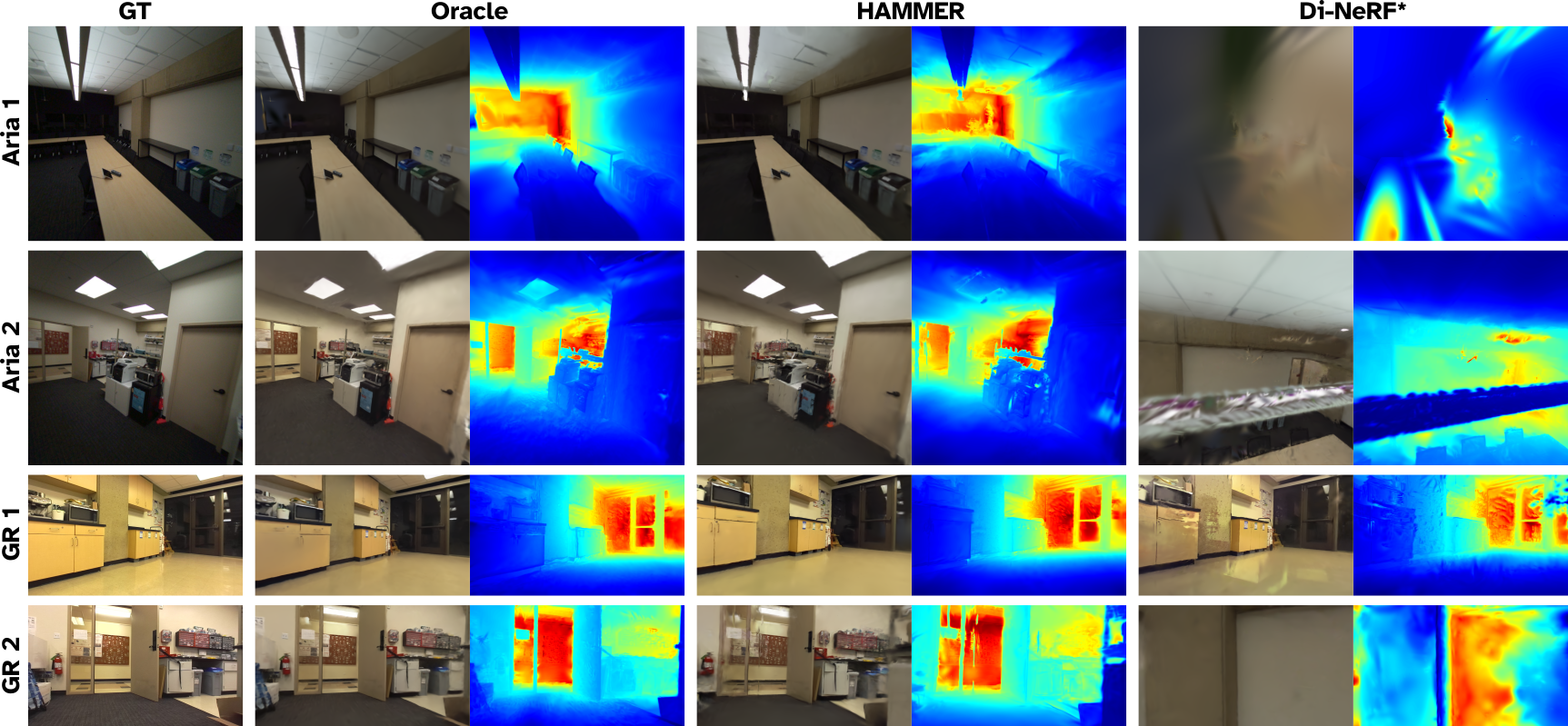}
    \caption{Rendered evaluation images RGB (left) and depth (right) of HAMMER and two baselines across devices. HAMMER is visually and geometrically superior to Di-NeRF*, and approaches the quality of the \emph{Oracle} upper bound baseline.}
    \label{fig:render}
    \vspace{-10pt}
\end{figure*}

\subsubsection{Scene Details}
The \emph{MultiAgentReplica} dataset contains four indoor scenes: \texttt{Off-0}, \texttt{Apt-0}, \texttt{Apt-1}, and \texttt{Apt-2}. Scenes have one to three rooms with no lighting variation.

Our hardware trials consist of two real-world scenes: \texttt{Home} and \texttt{Basement}. \texttt{Home} is a 6 room apartment and contains regions of high feature richness (open rooms with plants/furniture) and low richness (hallways and rooms with blank walls). The scene also includes regions of drastic lighting changes due to open windows with outside natural light, resulting in wide variations in color image appearances due to camera exposure compensation. We deploy two Aria glasses and one ground robot in this scene. \texttt{Basement} is composed of a large lab space, kitchenette, long hallways, and a conference room. This environment includes several regions with sparse visual features like hallways and reflective floor-to-ceiling glass windows. Here, two Aria glasses and two ground robots are deployed.

\subsubsection{Baselines}
In \emph{MultiAgentReplica}, we benchmark HAMMER against CP-SLAM \cite{hu2023cp} and MAGiC-SLAM \cite{yugay2024magic}. However, importantly, neither of the public implementations of the baselines are real-time capable. In the real-world trials, we compare HAMMER to three different baselines. \edit{The first is \emph{Oracle}, which uses bulk SfM on the same sensor data used by HAMMER to optimize a $3$DGS map.} \emph{Oracle} provides an upper bound on the performance of HAMMER because the bulk SfM poses serve as a ``ground truth" in our evaluation set. Next, \emph{Individuals} is a set of baselines, where each corresponds to a single robot optimizing its own independent $3$DGS maps from its own data. Because each robot cannot map the entire scene, any single map is not generalizable, highlighting the need for a global map. 

Because both MAGiC-SLAM and CP-SLAM are not real-time capable, do not have publicly available code for integration with hardware/ROS, and do not support heterogeneous devices,  we cannot compare these methods to HAMMER in the real-world trials. Instead we compare to Di-NeRF \cite{asadi2024di}, which proposes collaboratively optimizing a NeRF using gradient-based robot frame alignment. We modify Di-NeRF to provide a fair comparison to HAMMER by centralizing its map optimization (previously distributed), training it on the data cached during HAMMER's runtime, and replacing its NeRF with a $3$DGS. We refer to this baseline as \emph{Di-NeRF*}.

\subsection{Results}
\subsubsection{MultiAgentReplica} In Table \ref{tab:replica}, we compare the scene reconstructions performance of HAMMER to MAGiC-SLAM and CP-SLAM. HAMMER outperforms both baselines on all averaged metrics, and does so at least 25$\times$ faster than CP-SLAM and 16$\times$ faster than MAGiC-SLAM. The key advantage of HAMMER over these baselines is that it decouples local robot pose estimation from the joint mapping process, leading to significantly faster runtimes. Superior reconstruction results also demonstrate that this decoupling does not degrade scene reconstruction.

\subsubsection{Hardware Trials}
HAMMER is able to successfully construct quality $3$DGS maps, visualized in (Fig. \ref{fig:mesh}) as the 3DGS evolves in time during the online map optimization process. Embedded in the figure are the trajectories traversed by the robots, along with colors indicating where each robot contributed data. Qualitatively, HAMMER demonstrates superior novel-view renders on the evaluation set (Fig. \ref{fig:render}) compared to Di-NeRF*, which fails to resolve robot alignments and therefore cannot accurately match the ground-truth images. HAMMER also approaches the quality of \emph{Oracle}.

\begin{figure}[]
    \vskip 5pt
    \centering
    \includegraphics[width=0.9\linewidth, trim={0 0.5em 0 1.5em}, clip]{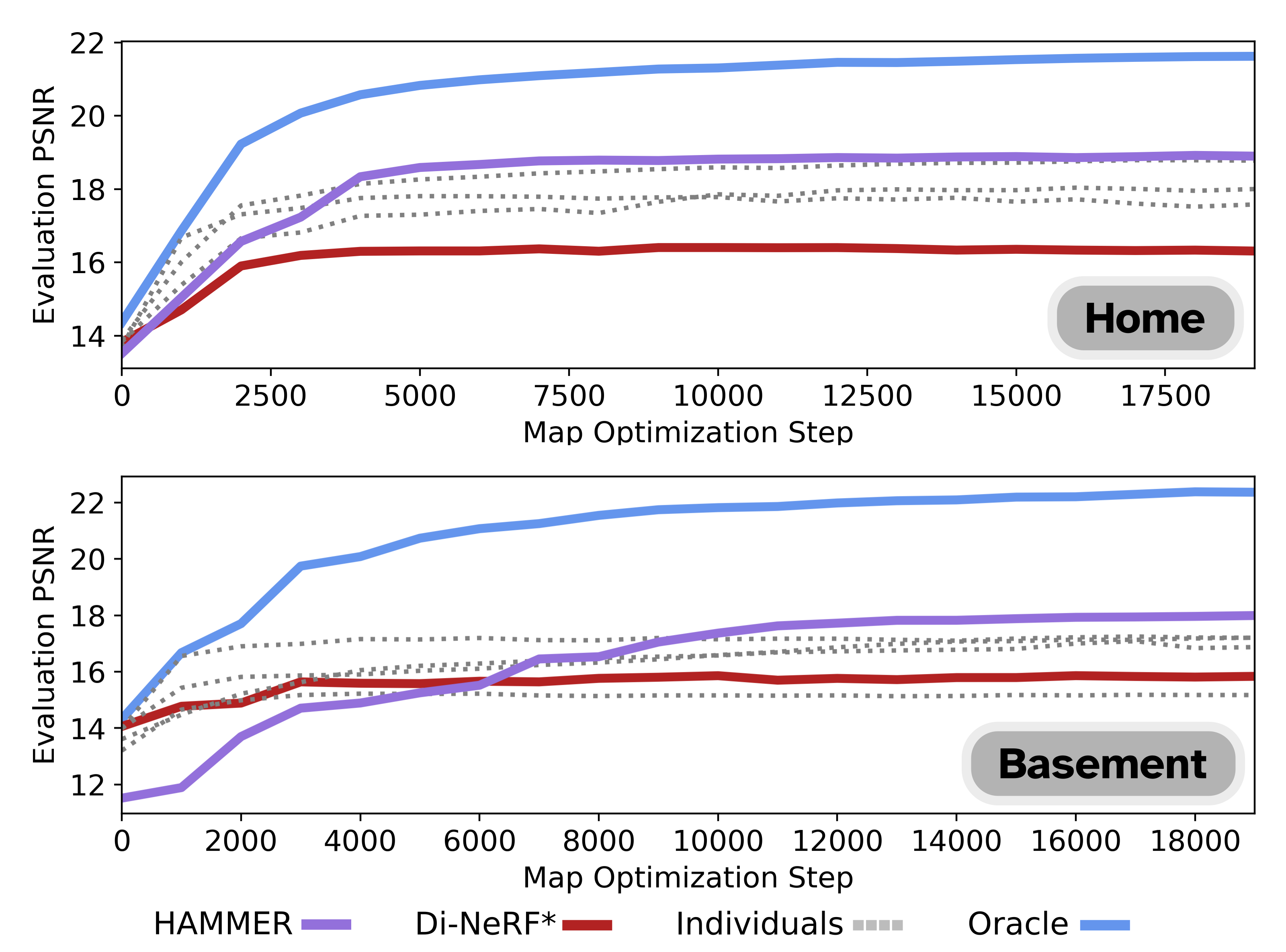}
    \caption{Map quality over time for HAMMER and baselines in two scenes. HAMMER outperforms \emph{Di-NeRF*}, demonstrating the necessity of accurate robot alignment. It also outperforms \emph{Individuals}, highlighting the benefits of collaboration. \emph{Oracle} serves as an upper bound on map quality.}
    \label{fig:psnr}
    \vspace{-10pt}
\end{figure}

We quantify reconstruction quality based on the average PSNR of each method on the held-out evaluation dataset  (Fig. \ref{fig:psnr}), comprised of 10 held-out frames per device. HAMMER dramatically outperforms Di-NeRF* which fails to converge to accurate inter-robot alignments. \edit{\emph{Individuals} underperforms HAMMER since evaluation views span the entire scene, highlighting the benefit of robot collaboration in map building. Note that the performance of some individual maps is relatively high due to good coverage of the entire scene from particular robots.} Finally, \emph{Oracle} upper bounds the performance of HAMMER \edit{due to access to ``ground-truth" poses. It also has access to the entire dataset at optimization time, leading to more optimal reconstructions compared to those generated from streaming data like in HAMMER. However, \emph{Oracle} is much slower to compute.}

\subsubsection{Runtime}
\label{sec:timing}
HAMMER runtime is dominated by the inter-robot alignment module (a one-time computation for each robot), and the $3$DGS map optimizer (an on-going computation to integrate streaming data into the map). The runtime of the alignment module is dominated (90\% - 99.5\% of the runtime) by the localized SfM procedure. Each SfM alignment requires about 33.6s to solve. The remaining runtime is primarily the feature-matching verification for proposed image correspondences (approximately 2.6s). \edit{In comparison, computing the batch COLMAP solution on all images for the \textit{Oracle} baseline takes between 2 and 5 hours depending on the configuration of COLMAP, illustrating the need for HAMMER's alignment module, which requires a one-time computation (per device) of about 36 sec.} For the continual $3$DGS map optimizer, processing a single data-frame ranges between 15 ms (66 Hz) when the map is sparsely populated to 38 ms (26 Hz). Slower iterations typically occur immediately after a robot has been aligned and all of its cached data is consolidated into the map. 

\edit{Bandwidth and communication overhead required by HAMMER are dependent on sensor configuration. Data streams for each device in our hardware experiments operate at 30-35 Mbps. Of that total bandwidth, the depth data is 25-30 Mbps and the compressed color images are approximately 2 Mbps. This bandwidth can be sustained by high-end commercial WiFi routers. Applying depth compression \cite{wilson2017fast} would substantially reduce the required bandwidth, although this technology was not directly supported by our devices. Down-sampling data in both transmission frequency and resolution could also reduce bandwidth overhead.}

 \begin{table}[]
    \centering
    \begin{tabular}{c l c c ccc}\hline 
        \textbf{Methods} & \textbf{Metrics} & \textbf{Off-0} & \textbf{Apt-0} & \textbf{Apt-1} & \textbf{Apt-2} & \textbf{Avg.} \\ \hline
        \textbf{CP-SLAM} & PSNR [dB] $\uparrow$ & 28.56 & 26.12 & 12.16 & 23.98 & 22.71 \\
        \cite{hu2023cp} & SSIM $\uparrow$ & 0.87 & 0.79 & 0.31 & 0.81 & 0.69 \\
        & LPIPS  $\downarrow$ & 0.29 & 0.41 & 0.97 & 0.39 & 0.51 \\
        & Depth L1 [cm] $\downarrow$ & 2.74 & 19.93 & 66.77 & 2.47 & 22.98 \\
        & Time [min] $\downarrow$ & 200+ & 200+ & 200+ & 200+ & 200+ \\ \hline 
        \textbf{MAGiC-SLAM} & PSNR [dB] $\uparrow$ & \secondplace38.53 & \secondplace35.87 & \secondplace28.57  & \secondplace30.01 & \secondplace33.24 \\
        \cite{yugay2024magic}& SSIM $\uparrow$ & \firstplace0.98 & \firstplace0.97 & \firstplace0.93 & \firstplace0.94 & \firstplace0.95 \\
        & LPIPS  $\downarrow$ & \secondplace0.07 & \secondplace0.21 & \secondplace0.17 & \secondplace0.19 & \secondplace0.16 \\
        & Depth L1 [cm] $\downarrow$ & \firstplace0.48 & \firstplace0.72 & \secondplace5.85 & \firstplace1.17 & \secondplace2.05 \\
        & Time [min] $\downarrow$ & \secondplace85 & \secondplace146 & \secondplace158 & \secondplace144 & \secondplace133 \\ \hline
        \textbf{HAMMER} & PSNR [dB] $\uparrow$ & \firstplace43.02& \firstplace40.89 & \firstplace31.43  & \firstplace33.78 & \firstplace37.28 \\
        & SSIM $\uparrow$ & \firstplace0.98 & \firstplace0.97 & \secondplace0.90 & \firstplace0.94 & \firstplace0.95 \\
        & LPIPS  $\downarrow$ & \firstplace0.04 & \firstplace0.06 & \firstplace0.13 & \firstplace0.11 & \firstplace0.09 \\
        & Depth L1 [cm] $\downarrow$ & \secondplace0.94 & \secondplace1.27 & \firstplace2.40 & \secondplace1.25 & \firstplace1.47 \\
        & Time [min] $\downarrow$ & \firstplace8 & \firstplace8 & \firstplace8 & \firstplace8 & \firstplace8 \\ \hline
    \end{tabular}
    \caption{Map reconstruction performance based on training view synthesis between HAMMER and other multi-robot mapping methods on \textit{ReplicaMultiAgent}. Best and second best values are highlighted in green and yellow, respectively. CP-SLAM metrics are drawn from \cite{yugay2024magic}, with time estimated based on hardware used to run other methods.}
    \vspace{-10pt}
    \label{tab:replica}
\end{table}
\subsubsection{Multi-robot Semantic Motion Planning}
\begin{figure}[]
    \centering
    \includegraphics[width=0.95\linewidth]{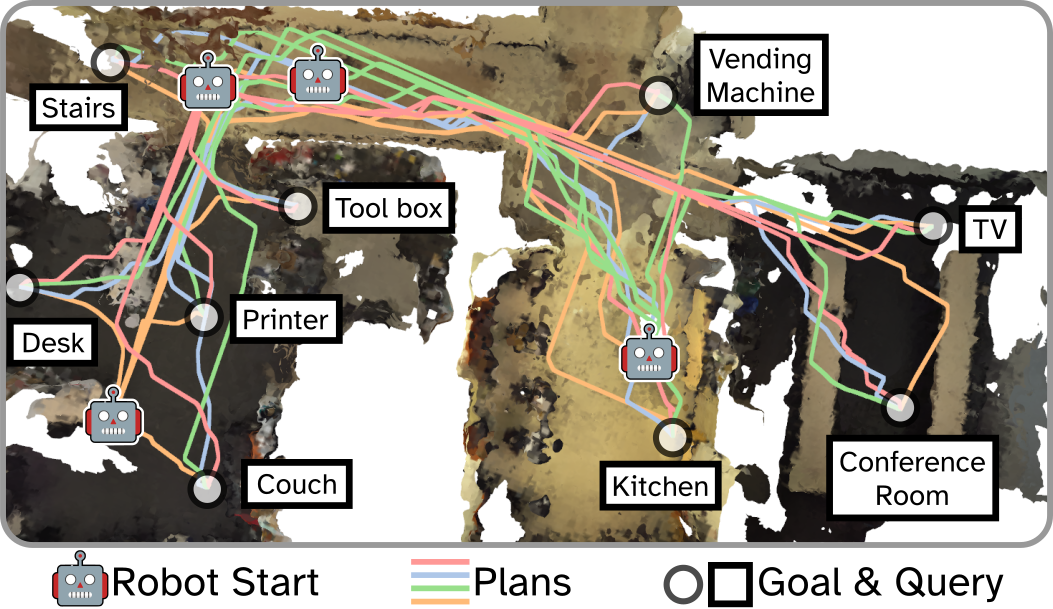}
    \caption{Motion plans for robots navigating to language-specified goals with trajectories from Splat-Nav \cite{chen2024splat}.}
    \label{fig:traj}
    \vspace{-15pt}
\end{figure}

As a brief qualitative case study and to highlight the utility of the semantic $3$DGS map for downstream tasks, we demonstrate that a map constructed online with HAMMER can be used for multi-robot motion planning task with language-specified goals in \texttt{Basement} \edit{after about 4 minutes of wall-clock time}. Goals are determined by searching the map for the Gaussians with the highest semantic similarity to the query (e.g. ``couch"). Once the goal is located, we use Splat-Nav \cite{chen2024splat} to plan a collision-free trajectory for each robot to the goal location. \Cref{fig:traj} shows optimized trajectories for each of the four robots in \texttt{Basement} with 9 semantically-specified goals. Goal location required about 20ms, and the subsequent trajectories were computed within 0.5s. All robots successfully completed the task.

\section{Conclusion and Limitations}
\label{sec:conc}
HAMMER has several limitations. First, devices without pose estimates cannot be used. Moreover, very noisy estimates or significant drift cannot be rectified by HAMMER's pose refinement, resulting in degraded maps.  Furthermore, the current implementation of HAMMER does not account for RGB-only devices, but we believe HAMMER can be extended to RGB-only by using monocular depth estimation. \edit{Finally, very short robot deployments render the one-time alignment cost prohibitively expensive.}

HAMMER is the first online multi-robot 3DGS mapping pipeline, built for heterogeneous robot teams. Maps produced with HAMMER can be used for a variety of downstream tasks like language-guided navigation. We show that HAMMER outperforms comparable baselines on real datasets and approaches the visual fidelity of an oracle in less than $10\%$ of the time. Future work will focus on better methods for data curation, volume reduction of redundant data, and detailed scene reconstructions over larger areas. \edit{The flexibility of HAMMER also facilitates map constructions from a wide range of robot embodiments (e.g. Aria glasses) and investigating the benefits of fusing diverse sets of devices is an interesting avenue for future research.}

\section*{Acknowledgment}
The authors would like to thank Dr. Ola Shorinwa for his valuable comments and thoughtful advice on this work.

\bibliographystyle{IEEEtran}
\bibliography{refs.bib}

\end{document}